%% file: iclr2026_conference.tex
\documentclass{article} 
\usepackage{iclr2026_conference,times}

\input{math_commands.tex}

\usepackage{hyperref}
\usepackage{url}
\usepackage{booktabs}
\usepackage{array}
\usepackage{tabularx}
\usepackage{hyperref}

\newcolumntype{Y}{>{\raggedright\arraybackslash}X}

\title{Position: Logical Soundness is not a Reliable Criterion for Neurosymbolic Fact-Checking with LLMs}

\author{Jason Chan, Robert Gaizauskas \& Zhixue Zhao \\
School of Computer Science\\
University of Sheffield\\
United Kingdom \\
\texttt{\{jlychan1,r.gaizauskas,zhixue.zhao\}@sheffield.ac.uk} \\
}

\iclrfinalcopy 
\begin{document}

\maketitle

\begin{abstract}
As large language models (LLMs) are increasing integrated into fact-checking pipelines, formal logic is often proposed as a rigorous means by which to mitigate bias, errors and hallucinations in these models' outputs. For example, some neurosymbolic systems verify claims by using LLMs to translate natural language into logical formulae and then checking whether the proposed claims are logically sound, i.e. whether they can be validly derived from premises that are verified to be true. \textbf{We argue that such approaches structurally fail to detect misleading claims due to systematic divergences between conclusions that are logically sound and inferences that humans typically make and accept.} Drawing on studies in cognitive science and pragmatics, we present a typology of cases in which logically sound conclusions systematically elicit human inferences that are unsupported by the underlying premises. Consequently, we advocate for a complementary approach: leveraging human-like reasoning tendencies of LLMs as a feature rather than a bug, and using these models to validate the outputs of formal components in neurosymbolic systems against potentially misleading conclusions.
\end{abstract}

\section{Introduction}

Large language models (LLMs) have enabled substantial progress in automated fact-checking and claim verification \citep{survey_on_fact_checking, Augenstein2024, huang2025unmaskingdigitalfalsehoodscomparative}. However, their outputs remain vulnerable to hallucinations, bias, and logical inconsistency \citep{hallucination_survey, Quelle2024-td, pelrine-etal-2023-towards}. To mitigate these limitations, a growing body of work has proposed neurosymbolic approaches that use formal logic to constrain or validate LLM outputs. In some architectures, for example, LLMs act as semantic parsers that translate natural-language statements, e.g. from user input or model-generated text into logical formulae such as in first-order logic (FOL) that are then evaluated against a verified knowledge base or external source \citep{wang-shu-2023-explainable,foldcover}. A common underlying assumption is that if a claim is logically sound, i.e. validly derivable from true premises in a verified knowledge base under formal inference rules such as those of FOL, then the claim is supported and acceptable.

We argue that existing work relying on this approach has a critical blind spot: \textbf{logical soundness is not a reliable criterion for detecting misleading information, because human inferences often diverge from formal logical operations such that logically sound conclusions can be \textit{systematically} misleading}. This yields three implications that challenge the current paradigm: (a) as opposed to using formal logic to validate LLM outputs, LLMs also have the potential of modeling human inference and detecting misleading conclusions otherwise unrecognized by systems relying on formal logic; (b) research on LLM reasoning should investigate these models' human-like tendencies not only as bugs (i.e., models committing human-like genuine errors and biases in logical tasks) but also as features (i.e., models predicting inferences that humans would typically accept despite being unsound in formal logic); and (c) if LLMs are to model human reasoning effectively, they should not be trained or optimized to conform strictly to formal logic in their reasoning processes.

\section{Misleading Information and Logic-Based Approaches to Fact-Checking}

In the context of recognizing and mitigating misinformation, fact-checking aims to identify not only explicitly false statements but also those that are technically true yet misleading \citep{chen2024can, survey_on_deeplearning_for_misinformation,survey_on_fact_checking, akhtar-etal-2023-multimodal}. In line with existing literature, we use ``misleading'' to mean factually correct content that nonetheless conveys an \textit{implied meaning} that is false or unsupported \citep{tang-etal-2025-missing_half_truths}. Within this paradigm, the starting point of our position is that ``implied meaning'' includes inferences that human readers would typically draw or accept. Consider the sentence ``John had a fight with Marilyn and decided to break up with her.'' As predicted by the theory of conjunction buttressing, readers would typically interpret this compound statement as (i) a temporal sequence (fight first, breakup decision second) and often (ii) a causal relation (the fight contributed to the decision) \citep{10.7551/mitpress/5526.001.0001}. Accordingly, if John had already decided to break up with Marilyn before the fight, the sentence should be considered misleading despite being literally compatible with the facts.

In parallel, existing research combines LLMs with explicit logic-based verification to improve faithfulness and robustness in fact-checking. Example methods include using LLMs to decompose complex natural language claims into formulae composed of FOL predicates whose truth values are individually verified by external knowledge sources \citep{wang-shu-2023-explainable, vladika-etal-2025-step,foldcover}. Other methods rely on natural logic, using LLMs to segment claims and evidence into corresponding text spans, and then relying on automated theorem-provers in natural logic to determine whether the claims are logically entailed by the evidence \citep{krishna-etal-2022-proofver, strong-etal-2024-zero}. More generally, this approach of using LLMs to translate natural language into logical formulae which are then checked with a symbolic solver has also been proposed to ensure that LLMs' own reasoning processes and resulting outputs are logically sound, i.e. based on true premises and derived in a logically valid manner \citep{pei-etal-2025-fover, zheng2025correctnessexposingllmgeneratedlogical}. Likewise, formal logic underpins other verification paradigms, including knowledge-graph-based methods \citep{opsahl-2024-fact, hao-wu-2025-fact} (e.g. to infer unstated relationships between entities in graph) and program-based approaches \citep{pan-etal-2023-fact}. The underlying assumption across these logic-based approaches is that if a claim can be derived in a logically valid manner from premises that are trusted or assumed to be true, then the claim itself should be treated as supported and verified by that system. 

\section{The divergence between human reasoning and formal logic}

Existing work in cognitive science has shown that humans accept and make inferences in ways that diverge from what is considered valid under formal logical systems \citep{JohnsonLaird2010, Khemlani2019, ragni2019meta}. For example, consider a premise in disjunctive form: ``\textit{A or B}''. According to the theory of mental models, humans represent a disjunctive statement as a conjunction of possibilities that hold in default to the contrary. This account explains why, on hearing that  ``\textit{A or B}'', humans tend to accept the inferences ``\textit{it is possible that A}'' and ``\textit{it is possible that B}'' as intuitive and correct, even though both these inferences are invalid (\textit{and therefore unsound regardless of whether the premises are true}) in standard modal logic that is introduced to reason with statements about possibilities and necessities \citep{Johnson-Laird2025}. 

We now illustrate why such divergences between human inference and formal logic are problematic for logic-based approaches in fact-checking. While we present an example statement in disjunctive form (``\textit{A or B}'') commonly found in official announcements\footnote{See e.g. ``\textit{[i]f you look at the Summary of Economic Projections, things are moving by just a tick \underline{\textbf{or}} even a semi tick between now and March}'' (\url{https://www.federalreserve.gov/mediacenter/files/FOMCpresconf20180613.pdf})} and news reports\footnote{See e.g. ``\textit{fire [...] may have spread through the plane and caused the explosion, \underline{\textbf{or}} the jet could have caught fire after colliding with an object on the ground}'' (\url{https://www.bbc.co.uk/news/articles/ckgky7djx5eo}); ``\textit{Sometimes that's human error, someone misconfiguring something somewhere, \underline{\textbf{or}} in extreme cases a cyber attack}'' (\url{https://www.bbc.co.uk/news/articles/cev1en9077ro})} (thus making them realistic targets for fact-checking pipelines), we also cover a broader range of other linguistically natural phenomena and statements e.g. involving conjunctions, conditionals and modals in Table \ref{tab:typology-misleading}.

Given a knowledge base with only one verified true premise S1:

\begin{quote}
    \textbf{S1}: Tariffs for France will go up 10\%.
\end{quote}

Suppose a logic-based fact-checking system is tasked with verifying the claim that:

\begin{quote}
    \textbf{S2}: Tariffs for either France or some other European countries will go up 10\%.
\end{quote}

Denoting S1 in propositional logic as A, the system can validly derive S2 from S1 by applying the inference rule of disjunction introduction ($A \vdash A \vee B$), hence verifying that S2 is a logically sound conclusion given S1.\footnote{To be clear, our main concern is not that a system might spontaneously use disjunction introduction to generate new facts or statements wholly unrelated to its input context. Rather, the problem arises when a fact-checking system applies this rule to formally derive and verify an \textit{existing} claim that is in disjunctive form.}

As explained above however, humans typically accept the inference that ``\textit{it is possible that B}'' as following from a statement that ``\textit{A or B}''. On this basis, given S2, humans typically accept S2-I:

\begin{quote}
    \textbf{S2-I}: It is possible that tariffs for some other European countries will go up 10\%.
\end{quote}

even though S2-I is not explicitly supported by the premise S1.

S2 should therefore be considered misleading: it implies S2-I, a claim with no basis in S1. Taken altogether, the example illustrates how a conclusion verified by a system to be logically sound can still be misleading because human inference does not track formal logical operations. Assuming the same verified premise S1, Table \ref{tab:typology-misleading} now presents examples (including S2 and S2-I as discussed) in which logically valid operations yield sound conclusions that invite ungrounded human inferences. Together, these examples support our central claim that, \textbf{despite a common assumption underlying neurosymbolic approaches using LLMs, logical soundness is not a reliable criterion for detecting misleading statements when fact-checking for misinformation}.

\begin{table}[t]
\centering
\fontsize{7.4pt}{9.1pt}\selectfont
\setlength{\tabcolsep}{4pt}
\renewcommand{\arraystretch}{1.}
\begin{tabularx}{\columnwidth}{>{\raggedright\arraybackslash}p{0.19\columnwidth} >{\hsize=0.95\hsize}Y >{\hsize=0.92\hsize}Y >{\hsize=1.13\hsize}Y}
\toprule
\textbf{Logically Valid Operation Applied to S1} & \textbf{Conclusion Derived by Applying Operation (Misleading)} & \textbf{What Humans Infer from Conclusion} & \textbf{Cognitive/Pragmatic Basis of Human Inference} \\
\midrule
Disjunction Introduction\newline($A \vdash A \vee B$) & \textbf{S2}: ``Tariffs for either France or some other European countries will go up 10\%.'' & \textbf{S2-I}: ``It is possible that tariffs for some other European countries will go up 10\%.'' & Humans interpret disjunctions as a conjunction of default possibilities: i.e. given ``A or B'', inferring the possibility that ``B'' \citep{Johnson-Laird2025}. \\
Conjunction Introduction\newline($A \vdash A \wedge A$) & \textbf{S3}: ``Tariffs for France will go up 10\%, and tariffs for France will go up 10\%.'' & \textbf{S3-I}: ``Tariffs will go up 10\% twice in a row.'' & Humans infer that the repetition refers to two separate events (by the Gricean Maxim of Quantity \citep{LogicandConversation}), happening sequentially in time (by conjunction buttressing \citep{10.7551/mitpress/5526.001.0001}). \\
Material Implication (Positive Paradox)\newline($A \vdash B \rightarrow A$) & \textbf{S4}: ``If interest rates do not decrease, tariffs for France will go up 10\%.'' & \textbf{S4-I}: ``If interest rates \textbf{do} decrease, tariffs will \textbf{not} go up 10\%.'' & Conditional perfection. Humans routinely interpret natural-language conditionals as biconditionals \citep{270fada2-bb25-3f14-bb9d-59c9555f0d64,HORN2000289}. \\
Material Implication (Vacuous Truth)\newline($A \vdash \neg A \rightarrow B$) & \textbf{S5}: ``If tariffs for France will not go up 10\%, our domestic market will be flooded with French goods.'' & \textbf{S5-I}: ``If tariffs for France \textbf{do} go up 10\%, our domestic market will \textbf{not} be flooded'' & As above \\
By Axioms B and T in Modal Logic\newline($A \vdash \diamond A$) & \textbf{S6}: ``It is possible that tariffs for France will go up 10\%.'' & \textbf{S6-I}: ``It is possible that tariffs for France will \textbf{not} go up 10\%.'' & Humans infer uncertainty from a statement about possibility, by Gricean Maxim of Quality \citep{LogicandConversation,JOHNSONLAIRD2019103950}\\
\bottomrule
\end{tabularx}
\caption{Examples of logically sound derivations that can induce ungrounded human inferences.}
\label{tab:typology-misleading}
\end{table}

\section{Implications: Rethinking the Role of LLMs in Neurosymbolic Fact-Checking Systems}

Given the above, we argue that the current paradigm of constraining and validating LLM outputs through formal logic captures only part of what reliability in fact-checking and misinformation detection requires. Rather, \textbf{future work should also explore the potential for LLMs to evaluate whether formally sound conclusions (as verified or generated by logic-based systems) are likely to mislead human readers.}

This direction is plausible as prior work has shown that LLMs can be trained to recognize certain pragmatic inferences routinely made by humans \citep{shisen-etal-2024-large, ma-etal-2025-pragmatics,sravanthi-etal-2025-understand}. For example, LLMs improved through reinforcement learning \citep{somayajula-etal-2025-improving} can recognize scalar implicatures with certain gradable adjectives (e.g. the statement that ``\textit{the coffee is warm}'' implying that the coffee is not ``\textit{hot}'' \citep{nizamani-etal-2024-siga}) and encode such implicatures in their hidden representations \citep{lin-etal-2024-probing}. In parallel, existing work has also found that LLMs display various human-like reasoning tendencies \citep{eisape-etal-2024-systematic, mondorf-plank-2024-comparing}, even though studies in this category commonly investigate these tendencies only as systematic error patterns and cognitive biases in logical reasoning tasks (i.e., mistakes that people themselves would recognize once corrected) \citep{10.1093/pnasnexus/pgae233, parmar-etal-2024-logicbench}. For example, \cite{10.1093/pnasnexus/pgae233} and \cite{eisape-etal-2024-systematic} respectively found that LLMs exhibit \textit{content effect}, misjudging logical validity based on factual plausibility of the conclusion, and \textit{figural effect}, being unduly affected by the ordering of words and terms in the premises.

One possible approach to capitalize on these human-like pragmatic capabilities and reasoning tendencies is to use LLMs for what we term ``\textit{claim expansion}''. Returning to the earlier example, when a system is tasked with fact-checking S2, an LLM can be prompted to generate a set of \textit{n} \textit{follow-on} inferences that human readers are likely to make based on S2, especially including those which are logically unsound but intuitively acceptable such as S2-I.\footnote{The size (\textit{n}) of this set of follow-on inferences can be flexible and dependent e.g. on the importance of the claim being fact-checked and computational budget limits.} The system would then verify this expanded set of inferences using formal logic, and assign the original claim S2 a soft score based on the proportion of follow-on inferences that are logically verifiable. For example, because S2-I cannot be logically derived from S1, S2's score would decrease accordingly and reflect its potentially misleading nature. In this framework, our proposed method effectively utilizes LLMs' human-like reasoning tendencies and cognitive biases as a valuable feature rather than a bug to be eliminated. 

That said, we recognize that LLMs used in such a capacity still inherit certain challenges in terms of model hallucination and social biases \citep{hallucination_survey,gallegos-etal-2024-bias,kalai2025languagemodelshallucinate}. For example, similar to issues observed when using LLMs to decompose claims into atomic subclaims for verification (see e.g. \citealp{hu-etal-2025-decomposition}), LLMs may hallucinate or generate irrelevant follow-on claims. Furthermore, while human-like reasoning tendencies are desirable for our use case, models may still exhibit specific social biases when expanding claims to be fact-checked in a way that prejudices minority groups \citep{10.1145/3630106.3658975,shrawgi-etal-2024-uncovering}. In both these respects, our approach remains dependent on existing mitigation methods, such as more rigorous data curation \citep{gunasekar2024textbooks} and test-time interventions \citep{li2023inferencetime,siddique-etal-2026-shifting}, to improve models’ fairness and factuality. Moreover, existing work shows that LLMs can be prompted to explicitly express their confidence levels in their own generated outputs, in a way that provides well-calibrated estimates of how likely these outputs are correct \citep{lin2022teachingmodelsexpressuncertainty,tian-etal-2023-just,xia-etal-2025-survey}. Our proposed approach could incorporate these uncertainty scores either by rejecting outputs (i.e. model-generated follow-on inferences) that fall below a certain confidence threshold, or proportionately discounting the impact of low-confidence outputs when computing the overall soft score for a particular claim being verified. 

Taken altogether, our proposed approach shows that LLMs’ human-like reasoning tendencies and pragmatic capabilities can serve as valuable features, complementing formal logic-based verification by modelling inferences that are logically unsound but typically considered intuitive and acceptable by humans. Importantly, for LLMs to serve effectively in such a capacity, we argue that these models' training should not enforce or encourage conformity to formal logic. Methods that reward formal logical consistency \citep{calanzone2025logically} or bias training data toward strictly logic-conforming patterns \citep{NEURIPS2024_8678da90,tan-etal-2025-enhancing-logical} risk suppressing the very pragmatic sensitivities needed for LLMs to be useful in detecting logically sound but misleading outputs.

\section{Alternative Views}

\textbf{AV: We can simply augment logical soundness with an auxiliary criterion X}

Requiring that claims be both logically sound whilst meeting another criterion could systematically result in unintended false-positives. This is because, as we have demonstrated in the previous section, acceptable human inferences are often logically unsound. If a knowledge base contains the statement that ``\textit{it is possible that the Fed will raise interest rates by 0.5\%}'', we would not want our claim verification system to reject, e.g., ``\textit{it is possible that the Fed will not raise interest rates by 0.5\%}'', which is otherwise an intuitive and acceptable inference \citep{JOHNSONLAIRD2019103950}. However, rejecting claims only if they are logically unsound \textbf{and} do not meet a criterion X is equally problematic. As shown above, such a system would systematically fail to detect misleading statements such as the ones in Table \ref{tab:typology-misleading}.

\textbf{AV: There is no need to focus on compound statements involving logical connectives because malicious actors are unlikely to use them.}
\textit{Such actors do not care about logical soundness and simple assertive claims (corresponding to atomic propositions in logic) are often more persuasive.}

As LLMs have the potential of producing and spreading misinformation at scale \citep{Ferrara_2024}, we should expect that these models could be optimized to bypass safeguards. In other words, even if human actors prefer direct assertions, automated systems can learn to generate formally sound compound claims that evade logic-based filters while still inducing misleading inferences.

\textbf{AV: We can simply expose logic derivation traces to users and let humans decide which conclusions to accept.}

This view assumes, first, that users will consistently inspect formal traces, and second, that they can reliably detect pragmatic misleadingness from those traces. In practice, many users may not be capable of performing this review especially at scale.

Moreover, when users delegate trace interpretation to LLMs, the problem reappears: once again models are depended upon to flag mismatches between formal logic and typical human inferences. As such, trace transparency does not eliminate the reliability issue raised here.

\section{Conclusion}

In this work, we argue that logical soundness is not a reliable criterion for assessing claims in the context of fact-checking and misinformation detection, as it \textit{systematically} fails to detect certain statements that are misleading due to the divergence between human reasoning and formal logic. We discuss and list examples of logically sound conclusions that are nonetheless misleading since as they tend to elicit ungrounded inferences from humans, as predicted by various cognitive science and linguistics studies. On this basis, we present an alternative to the current paradigm of constraining and validating LLM outputs with formal logic, and call instead for future work to explore and maximize the potential of LLMs for detecting such misleading statements that would otherwise be licensed in strictly formal systems.

\section*{Acknowledgment}

This work was supported by the UKRI AI Centre for Doctoral Training in Speech and Language Technologies (SLT) and their Applications funded by UK Research and Innovation [grant number EP/S023062/1]. For the purpose of open access, the author has applied a Creative Commons Attribution (CC BY) licence to any Author Accepted Manuscript version arising.

\bibliography{iclr2026_conference}
\bibliographystyle{iclr2026_conference}

\appendix

\end{document}

%% file: math_commands.tex
\usepackage{amsmath,amsfonts,bm}

\def\eqref#1{equation~\ref{#1}}

\def\1{\bm{1}}

\DeclareMathAlphabet{\mathsfit}{\encodingdefault}{\sfdefault}{m}{sl}
\SetMathAlphabet{\mathsfit}{bold}{\encodingdefault}{\sfdefault}{bx}{n}

